\documentclass[a4paper,10pt,twoside]{article}

\usepackage{clin}        
\usepackage{harvard}     
\usepackage{booktabs}
\usepackage{latexsym}
\usepackage{fancyvrb}
\usepackage{todonotes}
\usepackage{amssymb}
\usepackage{pifont}
\usepackage{color}
\usepackage[normalem]{ulem}

\newcommand{\cross}{\textcolor{red}{\ding{54}}}
\newcommand{\cmark}{\textcolor{green}{\ding{52}}}

\begin{document}

\title{Neural Semantic Parsing by Character-based Translation: Experiments with Abstract Meaning Representations}

\author{Rik van Noord$^*$ \email{r.i.k.van.noord@rug.nl}\\
{\normalsize \bf Johan Bos}$^*$ \email{johan.bos@rug.nl}\\
\AND \addr{$^*$Center for Language and Cognition Groningen (CLCG), University of Groningen, The Netherlands}} 

\newcommand{\notered}[1]{\todo[color=red!30,line,size=\tiny]{#1}}

\maketitle\thispagestyle{empty} 


\begin{abstract}
We evaluate the character-level translation method for neural semantic parsing on a large corpus of sentences annotated with Abstract Meaning Representations (AMRs). Using a sequence-to-sequence model, and some trivial preprocessing and postprocessing of AMRs, we obtain a baseline accuracy of 53.1 (F-score on AMR-triples). We examine five different approaches to improve this baseline result: (i) reordering AMR branches to match the word order of the input sentence increases performance to 58.3; (ii) adding part-of-speech tags (automatically produced) to the input shows improvement as well (57.2); (iii) So does the introduction of super characters (conflating frequent sequences of characters to a single character), reaching 57.4; (iv) optimizing the training process by using pre-training and averaging a set of models increases performance to 58.7; (v) adding silver-standard training data obtained by an off-the-shelf parser yields the biggest improvement, resulting in an F-score of 64.0. Combining all five techniques leads to an F-score of 71.0 on holdout data, which is state-of-the-art in AMR parsing. This is remarkable because of the relative simplicity of the approach.
\end{abstract}

\section{Introduction}

Various approaches to open-domain semantic parsing have been proposed in the last years. What we now could refer to as 
``traditional'' approaches are semantic parsers that use supervised learning to create a syntactic analysis on which the meaning representations are constructed, usually in a compositional way. Research in this area comprises \citeasnoun{BosClarkSteedman2004COLING},
\citeasnoun{mrs},
\citeasnoun{Butler:2010}, 
\citeasnoun{phongle},
\citeasnoun{lewis2014combined},
\citeasnoun{Bos2015NoDaLiDa},
\citeasnoun{artzi:15}, and many others.
Efforts to create datasets of sentences paired with meaning representations have stimulated research in semantic parsing \cite{amr,Bos2017GMB}, especially those using the formalism of Abstract Meaning Representation (AMR), for which also shared tasks have been organized \cite{sharedtask:16}. 
In this article, therefore, we concentrate
on semantic parsing of AMRs, because large gold-standard datasets are available and various different approaches can be compared.

In contrast to the traditional approaches mentioned above,
there have been interesting attempts recently to view semantic parsing as a translation task, mapping English expressions to logical forms under supervision of some deep learning method. 
\citeasnoun{dong2016language} used sequence-to-sequence (seq2seq) and sequence-to-tree (seq2tree) neural translation models to produce logical forms from sentences for four different datasets (but not AMRs).
\citeasnoun{riga:16} used a similar method to produce AMRs in the context of the previously mentioned shared task, but the performance of their neural parser was still far below the state-of-the-art. Despite this, their method inspired other researchers to adopt this seq2seq approach \cite{peng:17,konstas:17}. But, even though they got substantial improvements over \citeasnoun{riga:16}, their systems still did not come close to state-of-the-art. 
The neural approach of \citeasnoun{foland:17} did reach state-of-the-art performance, but they used five bi-LSTM networks instead of a single seq2seq model.

What all these attempts have in common, and why they are fascinating, is that they completely avoid complex models of the syntactic and semantic parsing process and therefore do not rely on heavily engineered features. However, except for \citeasnoun{riga:16}, they also only use word-level input. This is interesting, because \citeasnoun{riga:16} obtained a substantial improvement for their character-level model over their word-level model. 
Character-embeddings, 
since they were introduced by \citeasnoun{sutskever:11}, have also shown improvements in a number of areas, such as POS-tagging \cite{charpos:14,plank:16}, text classification \cite{zhang:15}, and, most importantly, Neural Machine Translation \cite{chung-cho-bengio:2016:WMT}.

The aim of this article is to find out how far we can push character-level neural semantic parsing: can we reach accuracy scores comparable with traditional approaches to semantic parsing? 
More specifically, our objectives are (1) try to reproduce the results of \citeasnoun{riga:16}; (2) improve on their results by employing several novel techniques; and (3) investigate whether injecting linguistic knowledge can improve neural semantic parsing.

We make three main contributions. First, we introduce novel techniques to improve neural AMR parsing. Second, we show that linguistic knowledge can still contribute to neural semantic parsing. Third, we show that adding silver standard to the training data makes a considerable (positive) difference in terms of performance. Our final model reaches an F-score of 71.0, which is the current state-of-the-art in AMR parsing.

\section{Method and Data}

We first give a bit of background on AMRs. Then we outline the basic ideas of the character-based translation model with English sentences as input and AMRs as output. We then establish a baseline system with the aim to improve it in the next section.

\subsection{Abstract Meaning Representations}

In our experiments utilizing neural semantic parsing we will focus on parsing Abstract Meaning Representations (AMRs). AMRs were introduced by \citeasnoun{Banarescu13abstractmeaning} and are acyclic, directed graphs that represent the meaning of a sentence. There are, in fact, three ways to display an AMR: as a graph, as a set of triples, or as a tree. An example of an AMR is shown in Figure~\ref{fig:example}, here displayed as a tree, the format that is used in the annotated corpora. The corresponding triple representation is shown in Table \ref{tab:triples}.

\begin{figure*}[!htb]
{
  \begin{minipage}{.7\textwidth}\centering
  \begin{Verbatim}[commandchars=\\\{\}]
(a / affect-01
   :ARG0 (w / wave-04
            :ARG1 (h2 / heat)
            :location (c / country :wiki "France" :name (n / name :op1 "France")))
   :ARG1 (p / person
            :ARG0-of (s / strike-02
                        :mod (h / hunger-01
                                :ARG0 p))))
\end{Verbatim}
  \end{minipage}
  \caption{\label{fig:example}AMR representing the meaning of \emph{Hunger strikers were affected by France's heat wave.}}
}
\end{figure*}

\begin{table}[!htb]
\centering
\caption{\label{tab:triples}The AMR of \emph{Hunger strikers were affected by France's heat wave.} displayed as the set of instance, attribute and relation triples.}
\begin{tabular}{lll}
\toprule
\textbf{Instance}        & \textbf{Attribute}  & \textbf{Relation} \\ \midrule
(instance, a, affect-01) & (TOP, a, affect-01) & (ARG0, a, w)      \\
(instance, w, wave-04)   & (wiki, c, France)   & (ARG1, a, p)      \\
(instance, h2, heat)     & (op1, n, France)    & (location, w, c)  \\
(instance,  c, country)  &                     & (ARG1, w, h2)     \\
(instance, n, name)      &                     & (name, c, n)      \\
(instance, p, person)    &                     & (ARG0, s, p)      \\
(instance, s, strike-02) &                     & (mod, s, h)       \\
(instance, h, hunger-01) &                     & (ARG0, h, p)     \\ \bottomrule
\end{tabular}
\end{table}

An AMR consists of concepts that are linked to variable names with a slash. In the example above we have that \texttt{a} is an instance of the concept \texttt{affect-01}, and \texttt{p} is an instance of the concept \texttt{person} (note that the names of the variables are not important). Concepts can be related to each other by using two-place predicates, which are indicated by a colon. So, the first \texttt{:ARG0} is an ordered relation between \texttt{a} and \texttt{w}. Inverse relations are denoted by the suffix \texttt{-of}. Note that, if one concept relates to more than one other concept (for instance, in the example above, the node \texttt{a} is related to \texttt{w} via \texttt{:ARG0}, and to \texttt{p} via \texttt{:ARG1}), the order of these relations within the AMR is not important. 

AMRs also allow for a re-occurrence of variables: the concept \texttt{person} with variable \texttt{p} stands in a relation with \texttt{affect-01} as well as with \texttt{hunger-01}. The brackets are important, because they signal which relations belong to which concepts (the spacing used in Figure~\ref{fig:example} is optional and is only used to increase readability). Some of the concepts have a number as suffix that indicate a specific word sense. AMRs also include proper name reference resolution by including a link to a wikipedia entry (wikification).

For evaluation purposes, AMRs are converted into triples. The triples of the AMR in Figure~\ref{fig:example} are shown in Table~\ref{tab:triples}. 
 
The accuracy of an AMR parser is computed by precision and recall on matching triples between gold standard AMRs and system-produced AMRs, using the \texttt{SMATCH} system \cite{smatch:2013}.

For the evaluation of our experiments we use the sentences annotated with AMRs from LDC release LDC2016E25\footnote{https://catalog.ldc.upenn.edu/LDC2017T10}, consisting of 36,521 training AMRs, 1,368 development AMRs and 1,371 test AMRs.

This release also includes the PropBank frameset and comes with pre-aligned AMRs and sentences.
In all results shown in this article, the models are trained on the training data. As development and test data we use the designated dev and test set from LDC2016E25, which are the exact same sets that are used in LDC2015E89. We remove HTML-tags from the input sentences, but URLs are kept in. 

\subsection{The Basic Translation Model}

To create our sequence-to-sequence translation model, we use the OpenNMT system \cite{2017opennmt}.  In contrast to \citeasnoun{peng:17} and \citeasnoun{konstas:17}, who use word-level input, we use character-level input .\footnote{We did experiment with word-based models, but they never obtained F-scores higher than 30.0. This is in line with \citeasnoun{peng:17} and \citeasnoun{konstas:17}, who only arrived at their final F-scores by applying extensive anonymization methods.}

We train a model with bidirectional encoding and general attention \cite{luong15}.
Since training a full model takes two to three days on a GPU, we perform a heuristic parameter search instead of an exhaustive one. We started out with a default model and changed only one parameter value in separate experiments. If we improved over the default, the setting was kept and combined with other parameter settings that improved performance. All models were only tested on the development set. Ultimately, we arrived at the settings shown in Table~\ref{tab:param}. All our described models in this paper are trained with these settings. Training is stopped 3 epochs after there is no improvement in validation perplexity on the development set anymore. The best performing model on the development set is then used to decode the test set.

\begin{table}[h]
\centering
\caption{Parameter settings of the seq2seq model.\label{tab:param}}
\begin{tabular}{ll|ll}
\toprule
\textbf{Parameter} & \textbf{Value} & \textbf{Parameter} & \textbf{Value} \\
\midrule
Layers             & 2              & RNN type           & brnn           \\
Nodes              & 500            & Dropout            & 0.3            \\
Epochs			   & 20--25			& Vocabulary		 & 100--200		  \\
Optimizer          & sgd        	& Max length         & 750            \\
Learning rate      & 0.1            & Beam size          & 5              \\
Decay              & 0.7            & Replace unk        & true           \\ 
\bottomrule
\end{tabular}
\end{table}

Following \citeasnoun{riga:16}, we do not want our model to learn the arbitrary characters that are used to represent variables. The characters itself do not carry any semantic information and are only necessary to indicate co-referring nodes. Therefore we remove all variables from the AMRs and simply duplicate co-referring nodes from the input. An example of such a preprocessed AMR is shown in Figure~\ref{fig:preprocessing}. Note that this means that we lose information, since the variables cannot be put back perfectly. We describe an approach to restore the co-referring nodes in the output in section 2.3.3. All wikification relations present in AMRs in the training set are also removed and restored in a post-processing step. Newlines present in an AMR are replaced by spaces, and multiple spaces are squeezed into single ones (so the input AMR is represented on a single line).\footnote{All pre- and post-processing scripts are available at https://github.com/RikVN/AMR}

\begin{figure*}[!htb]
{
  \begin{minipage}{.5\textwidth}\centering
  \begin{Verbatim}[commandchars=\\\{\}]
(m / material
   :mod (r / raw)
   :domain (o / opium)
   :ARG1-of (u / use-01
               :ARG2 (p / make-01
                        :ARG1 (h / heroin)
                        :ARG2 o)))
\end{Verbatim}
  \end{minipage}
  \begin{minipage}{0.5\textwidth}
  \begin{Verbatim}[commandchars=\\\{\}]
(material
   :mod (raw)
   :domain (opium)
   :ARG1-of (use-01
               :ARG2 (make-01
                         :ARG1 (heroin)
                         :ARG2 (opium))))
\end{Verbatim}
  \end{minipage}
  \caption{\label{fig:preprocessing}Example of the original AMR (left) and the variable-free AMR (right) displaying the meaning of \emph{Opium is the raw material used to make heroin.}}
}
\end{figure*}

\subsection{Postprocessing and Restoring Information}

The output of the seq2seq model is, of course, an AMR without variables, without wiki-links, and without co-occurrent variables.
Furthermore, because of the character-based seq2seq model, it could well be that there are brackets in the output that do not match, or that some nodes representing concepts are incomplete. This, obviously, needs to be fixed.

First, the variables in the AMRs are restored by assigning a unique variable to each concept. We also try to fix invalidly produced AMRs by applying a few heuristics, such as inserting parentheses and quotes, or by removing unfinished nodes. This is done by using the restoring script from \citeasnoun{riga:16}.\footnote{Taken from https://github.com/didzis/tensorflowAMR/} 
Then, we apply three methods to increase the quality of the AMRs. They are described below.

\subsubsection{Pruning}

A problem with many deep learning approaches is the fact that the decoder does not keep track of what it has already produced. As a consequence, we sometimes end up with duplicated, redundant material in our generated AMRs. This hurts precision. We propose four different methods to remove this redundant material. This is done on node level, where nodes are defined as relation-concept pairs without children, e.g. \texttt{:mod (raw)} and \texttt{:domain (opium)}.

\begin{table}[!htb]
\centering
\caption{\label{tab:pruning}Statistics of the different pruning methods. Methods were applied on the output of our baseline model on the dev set.}

\begin{tabular}{l|ccc}
\toprule
\textbf{Model} & \textbf{Nodes pruned} & \textbf{AMRs changed} & \textbf{F-score} \\ \midrule
Baseline                                                   & 0    & 0    & 54.8             \\ \midrule
Removing all re-occurrent nodes                            & 1426 & 689  & 55.4             \\
Removing re-occurrent nodes with same parent               & 135  & 95   & 55.0             \\
Removing re-occurrent nodes with frequency \textgreater 2  & 427  & 249  & 55.3             \\

\begin{tabular}[c]{@{}l@{}}Removing all re-occurrent nodes with same\\ parent, but also nodes with frequency \textgreater 2\end{tabular} & 496  & 302  & 55.5   \\ \bottomrule
\end{tabular}
\end{table}

The statistics of applying these four methods on our baseline model (dev set) are shown in Table~\ref{tab:pruning}. Note that all these processes are trade-offs: usually duplicates are correctly recognized as redundant and can be removed, but sometimes we erroneously remove actual re-occurrent nodes. 

The first method simply removes all re-occurrent nodes and is already quite effective: F-score increases by 0.6. The second method is more careful and only removes duplicate nodes if they have the same parent. This helps, but only by a small margin. The third method does not consider parent nodes, but removes nodes if they occur more than twice in the full AMR. This method also increases the F-score, but does not outperform the first method yet. The fourth method is a combination of the second and third method. All re-occurrent nodes with the same parent are removed, but also nodes occurring more than twice are removed. This results in the best F-score, an increase of 0.7 over the baseline. Two example AMRs whose branches are pruned using the fourth method are shown in Figure~\ref{fig:pruning}. 

\begin{figure*}[!htb]
{
  \begin{minipage}{.5\textwidth}\centering
  \begin{Verbatim}[commandchars=\\\{\}]
(material
       :mod (raw)
       \sout{:mod (raw)}
       :domain (opium)
       :ARG1-of (use-01
              :ARG2 (make-01
                     :ARG1 (heroin)
                     :ARG2 (opium))))
\end{Verbatim}
  \end{minipage}
  \begin{minipage}{0.5\textwidth}
  \begin{Verbatim}[commandchars=\\\{\}]
(material
       :mod (raw)
       :domain (opium)
              :mod (raw)
       :ARG1-of (use-01
              :ARG2 (make-01
                     :ARG1 (heroin)
                     \sout{:mod (raw)}
                     :ARG2 (opium))))
\end{Verbatim}
  \end{minipage}
  \caption{\label{fig:pruning}Example of pruned branches for the produced AMRs of \emph{Opium is the raw material used to make heroin.} In the left AMR, the second occurrence of \texttt{:mod (raw)} is already removed, because both branches are children of \texttt{material}. However, in the right AMR, none of the \texttt{:mod (raw)} branches share the same parent, so only the third occurrence is removed.}
}
\end{figure*}

\subsubsection{Wikification}

Since we removed wikification relations in preprocessing, our model will never output such a link. We restore wiki links in the output AMR by using an off-the-shelf system \cite{spotlight:2013}, following the method presented by \citeasnoun{bjerva2016meaning}. They look at the \texttt{:name} relations in an AMR and try to find this name on Wikipedia. If it has a page, the corresponding link gets added; otherwise the AMR remains unaltered. 

\subsubsection{Restoring co-referring nodes}

Our system also tries to restore co-referring nodes. If we output a duplicate node (a node already produced for this AMR), it replaces the node by the variable name of the node encountered first. This can only happen once per unique node, since the third instance of such a node is already removed in the pruning phase. An example of how the co-referring nodes are restored is shown in Figure \ref{fig:coref}.

\begin{figure*}[!htb]
{
  \begin{minipage}{.5\textwidth}\centering
  \begin{Verbatim}[commandchars=\\\{\}]
(m / material
       :mod (r / raw)
       :domain (o / opium)
              :mod (r2 / raw)
       :ARG1-of (u / use-01
              :ARG2 (m2 / make-01
                     :ARG1 (h / heroin)
                     :ARG2 (o2 / opium))))
\end{Verbatim}
  \end{minipage}
  \begin{minipage}{0.5\textwidth}
  \begin{Verbatim}[commandchars=\\\{\}]
(m / material
       :mod (r / raw)
       :domain (o / opium)
              \textbf{:mod r}
       :ARG1-of (u / use-01
              :ARG2 (m2 / make-01
                     :ARG1 (h / heroin)
                     \textbf{:ARG2 o})))
\end{Verbatim}
  \end{minipage}
  \caption{\label{fig:coref}Example of how co-referring nodes are restored. On the left an example of a produced AMR, on the right the AMR with co-reference restored.}
}
\end{figure*}

\subsection{Baseline Results}

Our first objective was to reproduce the results obtained by \citeasnoun{riga:16}. We did so, arriving at an F-score of 53.1 (see Table~\ref{tab:baseline}). Compared to the F-score of 43.0 by \citeasnoun{riga:16}, our score is significantly higher. This is probably due to the higher amount of training data and the fact that they used Tensorflow instead of OpenNMT. We also reproduced their results by using the exact same data, software and parameter settings as they did, obtaining an F1-score of 42.3.\footnote{We did not possess their Wikification and coreference restoring scripts, so differences might be attributed to that.}

\begin{table}[h]
\centering
\caption{\label{tab:baseline}Baseline Results Semantic Parsing on LDC2016E25.}
\begin{tabular}{ll|cc|cc}
\toprule
                          & \textbf{Type} & \textbf{Dev} & \textbf{Diff}  & \textbf{Test} & \textbf{Diff} \\ \midrule
\textbf{Baseline}         & seq2seq                      & 54.8  &         & 53.1    &           \\ \midrule
\textbf{Post-processing}  & Pruning                      & 55.5  &  + 0.7  & 53.7    & + 0.6     \\
                          & Restoring Co-reference       & 55.7  &  + 0.9  & 54.2    & + 1.1     \\
                          & Wikification                 & 55.8  &  + 1.0  & 54.1    & + 1.0     \\
\midrule
\textbf{All post-processing} & & 57.3 & +2.5 & 55.5  & + 2.4  \\
\bottomrule
\end{tabular}
\end{table}

As is shown in Table~\ref{tab:baseline}, concept pruning, restoring co-reference variables, and wikification all increase the F-score by about a percentage point each. This small gain of performance is what one could expect as each single operation has only a small impact on the overall contents of an AMR.

\section{Improving the Basic Translation Model} 

In the previous section we outlined our basic method of producing AMRs using a seq2seq model based on characters. In this section, we look at five different techniques to move beyond the F-score that we obtain with our basic method, that we will consider in this section as baseline. Some of the techniques were already (briefly) introduced in \citeasnoun{semeval:17}. 

\subsection{AMR Re-ordering} 

Although AMRs are unordered by definition, in our textual representation of the AMRs there is an order of the branches. However, these branches do not necessarily follow the word order in the corresponding English sentence. It has been shown that for (statistical) machine translation reordering improves translation quality \cite{collins2005clause}. We use the provided alignments to permute the AMR in such a way that it best matches the word order. We do this both on sub-tree level and on individual node level. The best matching AMR is defined as the AMR in which the order of the nodes (when traversing over the AMR depth-first) is the closest to the order of the words in the English sentence, following the alignments. An example of an AMR with a branch order best matching the input sentence is shown in Figure~\ref{fig:reshuffling}.

\begin{figure*}[!htb]
{
  \begin{minipage}{0.5\textwidth}
  \begin{Verbatim}[commandchars=\\\{\}]
(material
   :mod (raw)
   :domain (opium)
   :ARG1-of (use-01
               :ARG2 (make-01
                         :ARG1 (heroin)
                         :ARG2 (opium))))
\end{Verbatim}
  \end{minipage}
  \begin{minipage}{0.5\textwidth}
  \begin{Verbatim}[commandchars=\\\{\}]
(material
   :domain (opium)
   :mod (raw)
   :ARG1-of (use-01
               :ARG2 (make-01
                        :ARG2 (opium)
                        :ARG1 (heroin))))
\end{Verbatim}
  \end{minipage}
  \caption{\label{fig:reshuffling}Example of a variable-free AMR before (left) and after re-ordering (right) for the sentence \emph{Opium is the raw material used to make heroin}.}
}
\end{figure*}

We are also able to use this approach to augment the training data, since each reordering of the AMR provides us with a new AMR-sentence pair. Due to the exponential increase, large AMRs often have thousands of possible orders. We performed a number of experiments to find out how we could best exploit this surplus of data. Ultimately, we found that it is most beneficial to ``double'' the training data by adding the best matching AMR to the existing data set.\footnote{Instead of ordering the AMR nodes reflected by the word order of sentence, we also tried two different experiments based on consistency. The first experiment simply ordered the nodes alphabetically, without any other influence. This decreased the result of our baseline model by 2.0. Our second experiment was focused on fixing irregularities: if two nodes occur in a different order than they usually do (based on the full training set), we simply switch them around. This method did not change the order as considerably as the alphabetical ordering, but the result of the baseline model still decreased by 1.0. Hence we discarded both reordering techniques.}

\subsection{Introducing Super Characters}

We are not necessarily restricted to only using characters as input. For example, we can view the AMR relations (e.g. \texttt{:ARG0}, \texttt{:mod}) as atomic instead of a set of characters. This ensures that the characters for relations (e.g. \emph{m}, \emph{o} and \emph{d} for \texttt{:mod}) do not influence the general character embeddings of the concepts, which might improve performance. This way, we create a hybrid model that is a combination of word and character level input. An example of the AMR and sentence level input using super characters is shown in Figure~\ref{fig:superexample} and Figure~\ref{fig:posexample}.

\begin{figure}[htb]
\setlength{\arrayrulewidth}{0.01pt} 
\setlength{\tabcolsep}{1.2pt}
\begin{tabular}{p{3cm}*{29}{|c}|}
\toprule
\addlinespace
\cline{2-29}
AMR, chars: & ( & t & h & i & n & g & + & :      & q & u & a & n         & t & + & 1 & + & : & p & o & l & a & r & i & t & y & + & - & ) \\
\cline{2-29}
\addlinespace
\cline{2-16}
AMR, super chars: & ( & t & h & i & n & g & + & :quant & + & 1 & + & :polarity & + & - & )\\
\cline{2-16}
\addlinespace
\bottomrule
\end{tabular}
\setlength{\arrayrulewidth}{0.4pt}
\caption{\label{fig:supchar}\label{fig:superexample}
   Input for the AMR \texttt{(t / thing :quant 1 :polarity -)} representing the sentence \emph{Not one thing}, with and without super characters.
   The $+$-symbols represent spaces.}
\end{figure}

\begin{figure}[htb]
\setlength{\arrayrulewidth}{0.01pt} 
\setlength{\tabcolsep}{1.2pt}
\begin{tabular}{p{3cm}*{29}{|c}|}
\toprule
\addlinespace
\cline{2-21}
sentence: & I & +   & a & m & + & n   & o & t & + & t & h  & a & t & + & r & i & c  & h & + & .   \\
\cline{2-21}
\addlinespace
\cline{2-26}
sentence $+$ POS: & I & PRP & + & a & m & VBP & + & n & o & t & RB & + & t & h & a & t & IN & + & r & i & c & h & JJ & + & . \\
\cline{2-26}
\bottomrule
\end{tabular}
\setlength{\arrayrulewidth}{0.4pt}
\caption{\label{fig:posexample}Input for the sentence \emph{I am not that rich}, without and with POS-tags. POS-tags are inserted as super characters. The $+$-symbols represent spaces (word boundaries).}
\end{figure}

We also tried various ways to explicitly encode the tree structure by using super characters. In our basic model, the parentheses \emph{'('} and \emph{')'} are simply characters. This means that the model cannot differentiate between a parenthesis that opens the full AMR and a parenthesis that opens, say, the fifth subtree of the AMR. One would expect it would help the model if it has this information explicitly encoded in the input. For example, in an experiment we replaced each parenthesis in the structure by a super character that also provides the subtree information (e.g., an opening parenthesis on the fifth level becomes \texttt{*5*(}, while a closing bracket on the third level becomes \texttt{*3*)}. However this resulted in an F-score lower than the baseline and we discarded the technique.

\subsection{Adding Part-of-Speech Information}

We might still be able to benefit from syntactic information, even though we use a character-level neural semantic parser. To show this, we parse the sentences with the POS-tagger of the C\&C tools \cite{clark2003bootstrapping}, employing the Penn POS tagset. Each tag is represented as a single character and placed after the last character representation of the word that matches the tag (see Figure~\ref{fig:posexample}). Put differently, we create a new super character for each unique tag and add this to the input sentence. On the one hand, this will increase the size of the input. On the other hand, just a single character will add a lot of general, potentially useful, information. For example, proper nouns correlate with the \texttt{:name} relation, while adjectives correlate with the \texttt{:mod} relation.

\subsection{Adding Silver Standard Data}

A problem with neural parsing approaches is data sparsity, since a lot of manual effort is required to create gold standard data. \citeasnoun{peng:17} tried to overcome this by extensive generalization of the training data, but did not get near state-of-the-art results. \citeasnoun{konstas:17} applied a similar method, but also used the GigaWord corpus to self-train their system. They use their own pre-trained parser to parse the previously unseen sentences and add those to the training data in a series of iterations. Ultimately, their system is trained on 20 million additional data AMR-sentence pairs and obtains an F-score of $62.1$. Without this additional data, they obtain a score of $55.5$, which is better than \citeasnoun{peng:17}, but not close to state-of-the-art performance. 

Our method of obtaining new training data mainly differs from \citeasnoun{konstas:17} in two ways: 
(i) we use two off-the-shelf parsers to create the training data instead of self-training;
(ii) we employ a method to exclude lower-quality AMRs instead of using all available data. 
We therefore refer to this data as ``silver standard'' data, by which we mean something in between unchecked automatically produced data and gold standard data. 

Instead of self-training our parser, we use the off-the-shelf AMR parsers CAMR \cite{CAMR:15} and JAMR \cite{JAMR:14} to create silver standard data for our system. Both are non-neural, syntax-based parsers. CAMR works by first generating a dependency tree for the English sentence, after which it uses a transition-based algorithm to create the AMR graph. JAMR is the first published AMR parser and does the parsing in two stages: first identifying the concepts by using a semi-Markov model, and then identifying the relations between these concepts by searching for the maximum spanning connected subgraph.

Both systems are trained on the LDC2015E86 AMR corpus, which contains 16,833 training instances. We parse 1,303,419 sentences from the Groningen Meaning Bank \cite{gmb:lrec}, which mainly consists of newswire text. AMRs that are either invalid or include \texttt{null-tag} or \texttt{null-edge} (this is what the CAMR parser outputs when it is not able to find a suitable candidate parse) are removed.

We do not simply add the other AMRs to our data set. To ensure that the AMRs are at least of decent quality, we compare the produced AMRs with each other using \texttt{SMATCH} \cite{smatch:2013}. If their pairwise score does not exceed 55.0, the AMRs are not considered for adding to our training set. This value was picked to filter out AMRs that would only hurt the training process, but to also still include a large variety of AMRs and sentences. Our final set contained 530,450 sentences, that have both a CAMR and JAMR parse. 

We now have to determine which AMR to add to our silver data set. CAMR produces higher quality AMRs in general (64.0 vs 55.0 on the test set), but it might be beneficial to introduce some variety by also adding JAMR-parsed AMRs. We never add both CAMR and JAMR for the same sentence. We performed five experiments in which we added 100k silver AMRs, either containing 100\%, 75\%, 67\%, 50\% or 0\% CAMR-parsed AMRs. The results of testing on the development set are shown in Table~\ref{tab:camrjamr}. 

\begin{table}[htb]
\centering
\caption{\label{tab:camrjamr}F-scores on the dev set for adding different ratios of CAMR and JAMR parsed AMRs to our initial data set. All scores are without postprocessing improvement methods.}
\begin{tabular}{ccc}
\toprule
\textbf{\# CAMR AMRs} & \textbf{\# JAMR AMRs} & \textbf{F-score} \\ \midrule
100,000               & 0                     &   65.8           \\
75,000                & 25,000                &   65.8           \\
66,667                & 33,333                &   65.7           \\
50,000                & 50,000                &   65.3           \\
0                     & 100,000               &   61.4    		 \\ \bottomrule          
\end{tabular}
\end{table}

As would be expected, we see that only adding CAMR scores considerably better than only adding JAMR. However, the scores for adding 67\% and 75\% CAMR are very similar to adding 100\% CAMR. But, since this does not indicate that adding JAMR actually helps performance, we only add the CAMR-parsed AMRs in our silver data experiments. We randomly selected 20k, 50k, 75k, 100k and 500k instances for these experiments.

\subsection{Optimizing training}

Aside from the pre- and post-processing methods described, we can also optimize the training process itself. The first method we employ is pre-training on our full data set including silver AMRs, after which the model is fine-tuned on the gold data only. Both phases use the same parameter settings, as experiments with different learning rates resulted in lower performance. A similar procedure was used by \citeasnoun{konstas:17} and in general this is a method widely used in Neural Machine Translation \cite{denkowski:17}. 

The second method is averaging a set of models to decode the test set, instead of using a single model. This was first applied by \citeasnoun{averaging:16} as an alternative to the usual ensembling of models, which is known to give substantial improvements in Neural Machine Translation \cite{sutskever:14}. Ensembling, however, is very resource intensive, since the predictions of different models are averaged at decoding time. This as opposed to averaging, where the parameters of models are averaged to create a single model. This means that averaging, say, four models is four times faster than ensembling four models, while also using only a quarter of the memory the ensemble method uses. We tested with both ensembling and averaging and obtained similar results on the development set, thus opting to only use averaging in our experiments.

\section{Results and Discussion} 

Table~\ref{tab:resiso} shows the results of our improvement methods in isolation, meaning that only that individual method is added to our baseline model. Re-ordering has a clear positive effect, both for using the best re-ordering ($+2.0$) and adding that re-ordering to the existing data set ($+5.2$). Constructing super characters and adding POS-tags both lead to a similar increase in performance. Pre-training and subsequently fine-tuning also results in a substantial improvement, but creating an average model only has a slight positive effect. The biggest improvement comes from adding silver standard data to our training set, reaching a maximum of $65.8$ on the dev set. However, there is a limit with regards to adding silver data, since adding 500k silver AMRs performed worse than adding 50k, 75k or 100k silver AMRs. Finding the optimal number of silver AMRs is difficult due to the long training times and is therefore left for future work.

\begin{table}[!htb]
\centering
\caption{\label{tab:resiso}Results of the improvements in isolation (without post-processing).}
\begin{tabular}{ll|cc|cc}
\toprule
                         & \textbf{Type}  & \textbf{Dev} & \textbf{Diff} & \textbf{Test} & \textbf{Diff} \\ \midrule
\textbf{Baseline}        & seq2seq        & 54.8         &               & 53.1          &               \\ \midrule
\textbf{AMR Re-ordering} & Best           & 56.8         & + 2.0         & 55.1          & + 2.0         \\
                         & Doubling       & 60.0         & + 5.2         & 58.3          & + 5.2         \\ \midrule
 
\textbf{Introducing Super Characters} & Relations & 58.3 & + 3.5    & 57.4 & + 4.3    \\ \midrule

\textbf{Adding POS Tags}              & PTB      & 58.2 & + 3.4    & 57.2 & + 4.1    \\ \midrule

\textbf{Training optimization}        & Averaging        & 54.9 & + 0.1 & 53.4  & + 0.3   \\
                                      & Pre-training     & 59.4 & + 4.6 & 58.6  & + 5.5   \\
                                      & Both             & 59.5 & + 4.7 & 58.7  & + 5.6        \\ \midrule

\textbf{Adding Silver Standard Data}  & Adding 20k  & 62.2 & + 7.4  & 60.0 & + 6.9  \\
                                      & Adding 50k  & 64.7 & + 9.9  & 62.9 & + 9.8  \\
                                      & Adding 75k  & 65.7 & + 10.9 & 63.7 & + 10.6 \\
                                      & Adding 100k & 65.8 & + 11.0 & 64.0 & + 10.9 \\ 
                                      & Adding 500k & 63.8 & + 9.0  & 62.1 & + 9.0  \\ 
\bottomrule 
\end{tabular}
\end{table}

Since the previous experiments were all in isolation, we now test whether a combination of our methods still increases performance. The tested combinations are shown in Table~\ref{tab:rescheckmark}. Even after adding the silver data, the addition of POS-tags and super characters still increased the performance, albeit by a smaller margin. Interestingly, the best result (71.0) was not obtained by combining all improvement methods, since re-ordering the AMRs does not show an increase anymore after adding POS-tags and super characters. The best model without using any silver data obtains an F-score of 64.0, which is considerably higher than the AMR-only score (55.5) of \citeasnoun{konstas:17}. 

\begin{table}[!htb]
\centering
\caption{\label{tab:rescheckmark}F-scores for our neural models, combining the different improvement methods.}
\begin{tabular}{cccccc|cc}
\toprule
\multicolumn{1}{c}{\textbf{Post-proc}} & \multicolumn{1}{c}{\textbf{\begin{tabular}[c]{@{}c@{}}Adding\\ 100k Silver\end{tabular}}} & \multicolumn{1}{c}{\textbf{POS tags}} & \multicolumn{1}{c}{\textbf{\begin{tabular}[c]{@{}c@{}}Super\\ Chars\end{tabular}}} & \multicolumn{1}{c}{\textbf{\begin{tabular}[c]{@{}c@{}}Re-ordering\\ Best\end{tabular}}} & \multicolumn{1}{c}{\textbf{\begin{tabular}[c|]{@{}c@{}}Optimize\\ Training\end{tabular}}} & \multicolumn{1}{|c}{\textbf{Dev}} & \multicolumn{1}{c}{\textbf{Test}} \\
\midrule
\cross     & \cross            & \cross         & \cross       & \cross  & \cross          & 54.8  &  53.1          \\
\cmark     & \cross            & \cross         & \cross       & \cross  & \cross          & 57.3  &  55.5          \\
\cmark     & \cross            & \cmark         & \cmark       & \cmark  & \cmark          & 65.1  &  64.0              \\
\cmark     & \cmark            & \cross         & \cross       & \cross  & \cross          & 68.0  &  66.4          \\
\cmark     & \cmark            & \cmark         & \cross       & \cross  & \cross          & 68.9  &  67.3          \\
\cmark     & \cmark            & \cmark         & \cmark       & \cross  & \cross          & 70.4  &  69.0          \\
\cmark     & \cmark            & \cmark         & \cmark       & \cmark  & \cross          & 69.0  &  68.0          \\
\cmark     & \cmark            & \cmark         & \cmark       & \cross  & \cmark          & 71.9  &  71.0          \\
\bottomrule
\end{tabular}
\end{table}

Table~\ref{tab:others} shows the results of the most notable previous AMR parsing systems. Our best model outperforms all these previous parsers and reaches state-of-the-art results. However, we are also the first approach that uses the LDC2016E25 data set, which contains slightly more than double the number of gold standard training instances compared to the LDC2015E86 data set.\footnote{LDC2015E86 only contains 16,833 instances, as opposed to the 36,521 of LDC2016E25.} Therefore, we also trained the best performing model in Table \ref{tab:rescheckmark} on the LDC2015E86 data set, while still applying all our improvement methods. This model still obtains an F-score of 68.5, outperforming all previous AMR parsers, except for the parser of \citeasnoun{foland:17}.

\begin{table}[!htb]
\centering
\caption{\label{tab:others}F-scores for AMR parsing. Comparison with previously published results on the test set.}
\begin{tabular}{@{}l|l|c|c@{}}
\toprule
\textbf{Authors} & \textbf{Model}   & \textbf{Train set (gold)}    & \textbf{F-score} \\
\midrule
\citeasnoun{JAMR:14} & JAMR-14                      & LDC2013E117  & 58.0    \\
\citeasnoun{damonte:17} & AMR-eager					& LDC2015E86   & 64.0	 \\
\citeasnoun{artzi:15} & CCG parsing		            & LDC2014T12   & 66.3    \\
\citeasnoun{CAMR:15} & CAMR		                    & LDC2015E86   & 66.5    \\
\citeasnoun{JAMR:16} & JAMR-16                      & LDC2015E86   & 67.0    \\
\citeasnoun{pust:15} & SBMT	                        & LDC2015E86   & 67.1    \\
\midrule
\citeasnoun{riga:16}    & char-based seq2seq        & LDC2015E86   & 43.0    \\
\citeasnoun{peng:17}    & word-based seq2seq        & LDC2015E86   & 52.0    \\
\citeasnoun{konstas:17} & word-based seq2seq	    & LDC2015E86   & 55.5    \\
\citeasnoun{konstas:17} & word-based seq2seq + giga & LDC2015E86   & 62.1    \\
\citeasnoun{foland:17}  & 5 bi-LSTM networks (word-based) & LDC2015E86   & 70.7    \\
\midrule
This article & char-based seq2seq model + silver    & LDC2015E86    & 68.5   \\ 
This article & char-based seq2seq model + silver    & LDC2016E25    & 71.0   \\ 
\bottomrule                      
\end{tabular}
\end{table}

\citeasnoun{damonte:17} presented a way to evaluate system output in a more detailed way, by focussing on various aspects that are present in an AMR: the role labelling, word sense disambiguation, named entity recognition, wikification, detecting negation, and so on.
These detailed results of our best system are shown in Table~\ref{tab:detailedres}, in which
the results of the other parsers are taken from \citeasnoun{damonte:17}. Unfortunately, \citeasnoun{foland:17} did not publish these specific scores. As the table shows, our system scores higher than the other parsers on five of the eight metrics other than Smatch. In general, our system is quite conservative, obtaining a higher precision than recall for each metric.
Given the results in Table~\ref{tab:detailedres}, one would think that detecting negation and reentrancy would be ways to get an improvement in accuracy. Note that the other parsers score also relatively bad at these metrics. Compared to the other systems, our system scores worse on concepts, named entities, and wikification. A possible method to increase performance in the first two of those metrics is to adopt an anonymization or generalization approach for named entities and concepts, similar to \citeasnoun{peng:17} or \citeasnoun{konstas:17}.

\begin{table}[!htb]
\centering
\caption{\label{tab:detailedres}Comparison with previous parsers using the evaluation script of \protect\citeasnoun{damonte:17}. We also included precision and recall scores for our system.}
\begin{tabular}{@{}l|ccc|ccc@{}}
\toprule
                  & \textbf{CAMR} & \textbf{JAMR-16} & \textbf{AMR-eager} & \multicolumn{3}{c}{\textbf{Our system}}      \\ \midrule
\textbf{Metric}   & \textbf{F}    & \textbf{F}       & \textbf{F}         & \textbf{Pr.}  & \textbf{Rec.}  & \textbf{F}  \\ \midrule
Smatch            & 63            & 67               & 64                 & 76            & 67             & 71   \\
Unlabeled         & 69            & 69               & 69                 & 79            & 70             & \textbf{74}    \\
No WSD            & 64            & 68               & 65                 & 76            & 67             & \textbf{72}   \\
Reentrancy        & 41            & 42               & 41                 & 57           & 48              & \textbf{52}  \\
Concepts          & 80            & \textbf{83}      & \textbf{83}        & 87           & 78              & 82   \\
Named entities    & 75            & 79               & \textbf{83}        & 83           & 76              & 79   \\
Wikification      & 0             & \textbf{75}      & 64                 & 82           & 54              & 65  \\
Negations         & 18            & 45               & 48                 & 67           & 58              & \textbf{62}   \\
SRL               & 60            & 60               & 56                 & 70           & 62              & \textbf{66}   \\ \bottomrule
\end{tabular}
\end{table}

\section{Conclusion and Future Work}

Applying re-ordering of AMR branches, introducing super characters, and adding POS-tags are techniques that substantially improve neural AMR parsing using a character-based seq2seq model. However, the biggest increase of performance is triggered by adding a large quantity of silver standard AMRs produced by existing (traditional) parsers. This is in line with the findings of \citeasnoun{konstas:17}, who used the Gigaword corpus to get extra training data, although their training method is different from ours.

The obtained results are promising. 
Our best model, with an F-score of 71.0, outperformed any known previously published result on AMR parsing. This is remarkable, for traditional approaches are often based on extensive, manually crafted lexicons using linguistic knowledge. It should be noted, of course, that we use some linguistic knowledge in the form of POS-tags in our best models, and that we employ existing parsers trained on extensive linguistics annotations. In fact, one could consider the use of silver standard AMR data as a disadvantage, as there is still a need of an existing high-quality AMR parser to get the silver data in the first place. In our approach we rely even on two different off-the-shelf parsers. It would therefore be interesting to explore other opportunities, such as self-learning, as proposed by \citeasnoun{konstas:17}.

We have the feeling that there are still a lot of techniques that one could try to increase the performance of neural AMR parsing. From a more esthetical perspective, it would be nice if one could eliminate the AMR repair strategies that are used to resolve unbalanced brackets. An interesting candidate that could master this problem would be the seq2tree model presented by \citeasnoun{dong2016language}. Similarly, a more principled approach to deal with co-occurring variables would be desirable.

Another possible next step in semantic parsing is to change the target meaning representation. AMRs are unscoped meaning representations, and have no quantifiers. It would be challenging to transfer the techniques of neural semantic parsing to scoped meaning representations, such as those used in the Groningen Meaning Bank \cite{gmb:lrec} or the Parallel Meaning Bank \cite{eacl:pmb}.

\subsection*{Acknowledgements}

First of all we would like to thank Antonio Toral and Lasha Abzianidze for helpful discussion on neural AMR parsing and machine translation.
We thank the three anonymous reviewers for their comments. 
We would also like to thank the Center for  Information  Technology  of  the  University  of Groningen for their support and for providing access to the Peregrine high performance computing cluster. We also used a Tesla K40 GPU, which was kindly donated to us by the NVIDIA Corporation. This work was funded by the NWO-VICI grant ``Lost in Translation – Found in Meaning'' (288-89-003).

\bibliographystyle{clin} 
\bibliography{bibliography}  

\end{document}